\documentclass[conference]{IEEEtran}
\usepackage{times}

\usepackage[numbers]{natbib}
\usepackage{multicol}
\usepackage[bookmarks=true]{hyperref}
\usepackage{xcolor}
\usepackage{caption}
\usepackage{booktabs}
\usepackage{amsmath}
\usepackage{amssymb}
\usepackage{color, soul} 
\usepackage{graphicx}
\usepackage{makecell}
\usepackage{multirow}
\usepackage{colortbl}
\usepackage{float}

\newcommand{\tbl}[1]{Table~\ref{#1}}

\newcommand{\eg}[0]{{\em e.g.,~}} 
\newcommand{\ie}[0]{{\em i.e.,~}} 

\newcommand{\suppref}[1]{Appendix~#1}

\definecolor{idblue}{RGB}{90,155,213}
\definecolor{oodorange}{RGB}{247,150,70}

\begin{document}

\title{GHOST: Hierarchical Sub-Goal Policies for Generalizing Robot Manipulation}


\author{\authorblockN{Sriram Krishna\textsuperscript{1},
Ben Eisner\textsuperscript{1},
Haotian Zhan\textsuperscript{1},
Ying Yuan\textsuperscript{1},
Haoyu Zhen\textsuperscript{2}, \\
Chuang Gan\textsuperscript{2},
Shubham Tulsiani\textsuperscript{1},
David Held\textsuperscript{1}}
\authorblockA{\textsuperscript{1}Robotics Institute, Carnegie Mellon University\\}
\authorblockA{\textsuperscript{2}UMass Amherst\\}
{\Large\color{orange}\href{https://ghost-human-demo.github.io/}{\texttt{https://ghost-human-demo.github.io/}}}
}

\makeatletter
\let\@oldmaketitle\@maketitle
\renewcommand{\@maketitle}{\@oldmaketitle
\centering
  \begin{minipage}{0.85\linewidth}
    \centering
    \includegraphics[width=0.95\linewidth]{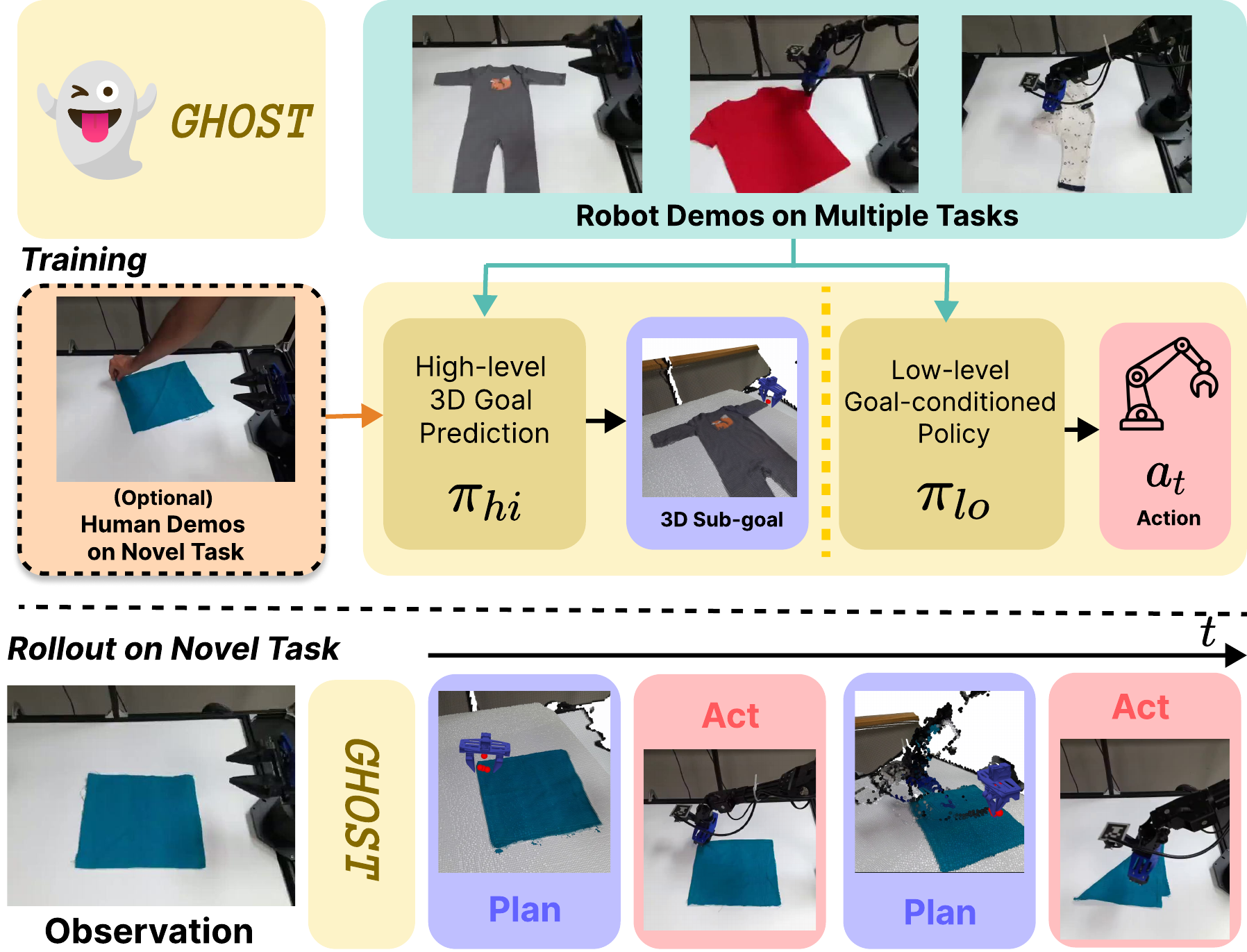}
  \end{minipage}
  \small
  \captionof{figure}{GHOST learns skills from robot teleoperation data and optionally uses human demonstrations to generalize to novel tasks. We train a hierarchical policy that decouples embodiment-agnostic goal prediction ($\pi_{hi}$) from embodiment-specific action execution ($\pi_{lo}$). By training $\pi_{hi}$ on both robot and human data and $\pi_{lo}$ purely on robot data, GHOST transfers learned manipulation skills to out-of-distribution tasks with third person human video demonstrations.}
  \label{fig:teaser}
}
\makeatother
\maketitle
\addtocounter{figure}{-1}

\begin{abstract}
We present \textbf{GHOST}, a framework for learning visuomotor manipulation policies that \emph{generalize} beyond the training distribution. GHOST factorizes control into (i) a high-level policy that predicts the next sub-goal as a \emph{distribution} over 3D end-effector poses from multi-view RGB-D observations, and (ii) a low-level goal-conditioned controller that executes embodiment-specific actions. To condition image-based policies on 3D goals, we introduce a simple spatial interface that projects predicted goals into the image plane and represents them as \emph{end-effector heatmaps}. Across a suite of manipulation tasks, this hierarchical factorization consistently improves performance and robustness compared to a flat Diffusion Policy. 

Further, we show that this hierarchical interface also makes it easy to incorporate human demonstrations without relying on (noisy) action retargeting. As sub-goals are largely embodiment-agnostic, we train the high-level policy on human video to specify how learned skills should be applied and composed, while keeping the low-level policy  trained purely on robot data. This hierarchy enables adaptation to novel objects and task variations using a small number of human demonstrations. 

\end{abstract}

\IEEEpeerreviewmaketitle

\section{Introduction}
\label{sec:introduction}

The dominant paradigm today for teaching robot policies to manipulate their environment is imitation learning (IL). Given a sufficiently large number of demonstration trajectories, imitation learning algorithms can learn precise, complex behaviors \cite{chi2025diffusion, zhao2023learning}. However, to generalize to novel objects and environments, they require massive amounts of data, typically collected through \textit{teleoperation}, where a human operator controls a robot to complete a given task \cite{black2024pi_0, khazatsky2024droid, lbmtri2025}. This recipe, while simple, is expensive and labor-intensive to scale. 

In this work we focus on an alternate lever: \textit{hierarchy}. Many manipulation tasks are naturally organized into phases - \eg a pick-and-place task may be decomposed into grasp, transport, and place phases. 
We argue that explicitly modeling this hierarchy improves robustness and data efficiency by separating \emph{sub-goal selection} from \emph{low-level execution}. 
Concretely, we define sub-goals as end-effector poses at phase boundaries, which provide a compact interface for composing skills and extending behavior over long horizons.

Based on this perspective, we propose a hierarchical policy that factorizes control into two modules. 
(i) A high-level policy $\pi_{hi}$ predicts the next sub-goal from visual observations and a language instruction, and
(ii) a low-level policy $\pi_{lo}$ conditioned on the predicted goal to produce embodiment-specific actions.
To connect these modules while retaining the benefits of image-based policies, we introduce a spatial goal representation that projects 3D sub-goals into the image plane and represents them as \emph{end-effector heatmaps}, enabling effective goal-conditioning of image-based policies.

This hierarchy yields two practical benefits. 
First, even when trained only on robot demonstrations, our hierarchical factorization significantly improves success rates compared to a flat Diffusion Policy \cite{chi2025diffusion}. 
Second, our insight is that this sub-goal interface is largely embodiment-agnostic, making it a convenient point to incorporate additional supervision. As a secondary contribution, we show that we can use human videos to train the high-level policy for novel task variants, without requiring retargeted action labels, enabling out-of-distribution generalization.

Thus, we present \textbf{GHOST}: \textbf{\underline{G}}eneralizing manipulation via \textbf{\underline{H}}ierarchical end-effect\textbf{\underline{O}}r \textbf{\underline{S}}ub-goals for Skill \textbf{\underline{T}}ransfer. GHOST is a hierarchical framework for training visuomotor policies that can generalize skills through human demonstrations. GHOST features: a) a 3D high-level goal prediction network that takes in RGB-D images to predict a distribution of sub-goal end-effector poses, and b) a low-level policy conditioned on goals represented as heatmap projections of the predicted end-effector sub-goals.

Our contributions are as follows:
\begin{itemize}
    \item A hierarchical framework for training policies with 3D end-effector keypoint sub-goals and heatmap-based goal-conditioning that consistently improves in-distribution performance over flat policies.
    \item A demonstration that this factorization naturally enables cross-embodiment transfer from human video without action retargeting, by training $\pi_{hi}$ on heterogeneous human and robot data while keeping $\pi_{lo}$ grounded in robot demonstrations.
\end{itemize}

\begin{figure*}[h]
\centering
\includegraphics[width=0.95\linewidth]{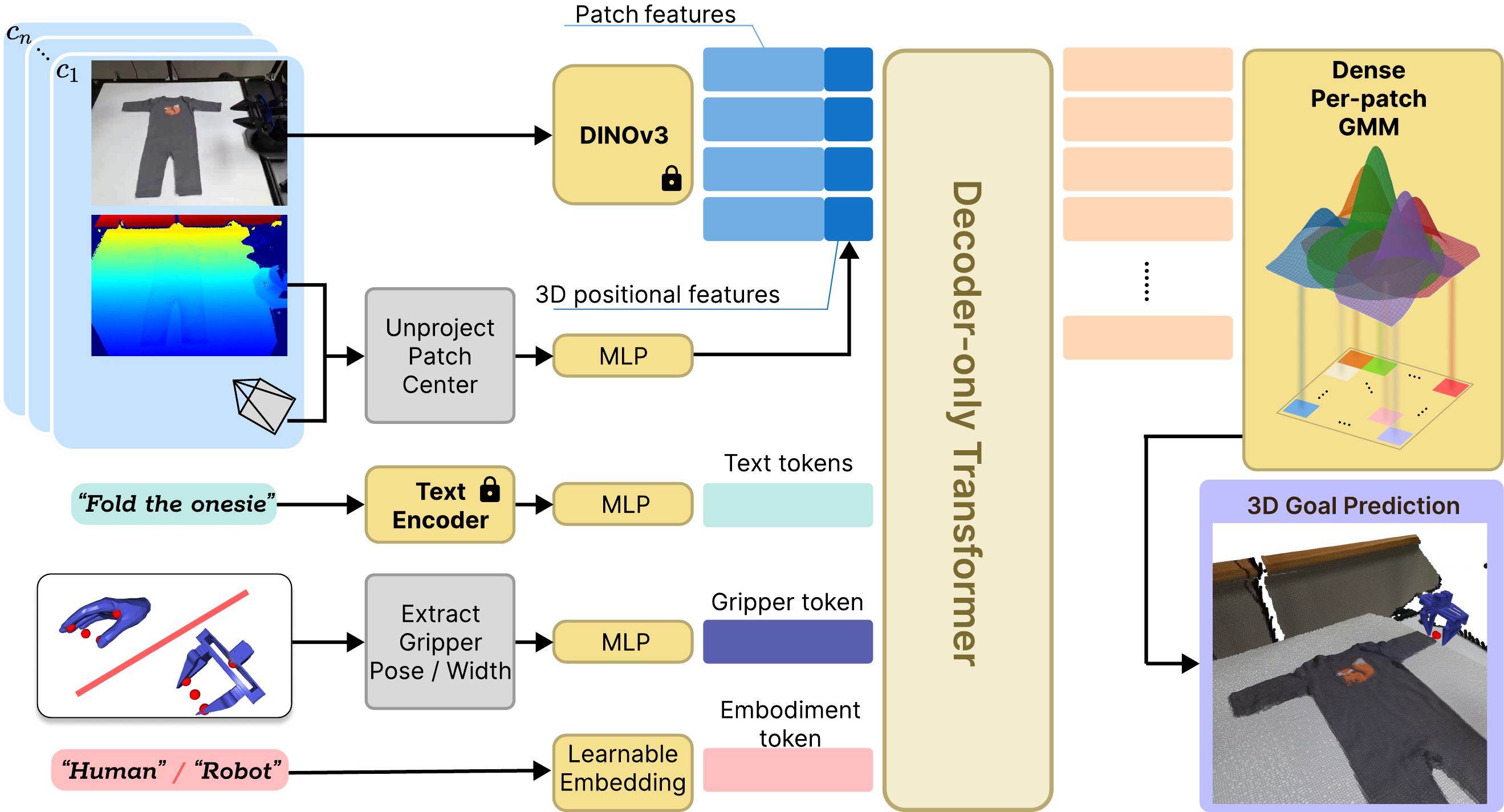}
\caption{\textbf{GHOST} High-level sub-goal prediction architecture: RGB-D observations from multiple cameras are processed with a DINOv3 encoder, with patch tokens augmented by 3D coordinates. Additional context (gripper state, language embedding, embodiment name) is encoded via separate MLPs. A decoder-only transformer processes all tokens, with each patch predicting the GMM parameters of a 3D sub-goal distribution over end-effector keypoints.}
\label{fig:high-level-arch}
\end{figure*}

\section{Related Work}
\label{sec:related_work}

\subsection{Hierarchical Imitation Learning} 
Hierarchical Imitation Learning factorizes the policy into a high-level ``planner'' and a low-level ``control'' policy. Previous work has primarily focused on learning high- and low-level policies from robot data \cite{mandlekar2020learning, xu2018neural, lynch2020learning}. 
This structure is particularly attractive for manipulation, where tasks often decompose into phases and where long-horizon behaviors can be expressed as sequences of sub-goals. 
Recent work has shown that representing sub-goals directly in end-effector space can yield strong generalization for articulated manipulation \cite{wang2025articubot}. 
In parallel, goal-conditioning has been explored through images and language, often by predicting intermediate targets and conditioning a controller on them \cite{bharadhwaj2024track2act, wen2023any, collins2025amplify}. Our work is most closely related to hierarchical approaches that use explicit spatial sub-goals. 
We differ in two ways: (i) we predict 3D end-effector sub-goals from multi-view RGB-D, and (ii) we introduce an \emph{end-effector heatmap} interface that makes goal-conditioning compatible with image-based policies.

\subsection{Learning from human demonstrations}
Large-scale robot teleoperation datasets have enabled impressive generalization in visuomotor policies \cite{khazatsky2024droid, o2024open, brohan2022rt, black2024pi_0}. 
However, collecting robot demonstrations remains expensive, motivating complementary sources of supervision such as human video \cite{goyal2017something, grauman2022ego4d, lepert2025phantom}. 
One line of work attempts to directly learn robot actions from human demonstrations by estimating hand pose and retargeting to a robot end-effector \cite{pavlakos2024reconstructing, potamias2025wilor, kareer2025egomimic, ren2025motion, haldar2025point, lepert2025phantom}. 
These approaches can suffer from noisy pose estimates and an embodiment mismatch. By restricting human supervision to embodiment-agnostic sub-goals, we avoid action retargeting and keep low-level control grounded in robot demonstrations.
Another line of work learns higher-level representations or latent dynamics from video corpora and maps them to robot actions with additional robot data \cite{ye2024latent, wang2023mimicplay}. Most similar to our work is MimicPlay~\cite{wang2023mimicplay}, which learns a latent distribution over hand trajectories collected through play data.  However, this approach relies on a video prompt of the task being provided at test-time, whereas we only require a language goal. In contrast, we treat human demonstrations as \emph{optional} supervision for the high-level planner in a hierarchy.

\section{Problem Statement and Assumptions} 
\label{sec:problem_statement}

We consider manipulation tasks that can be naturally decomposed into a sequence of sub-goal states $\{g_i\}_{i\in\left[ 1 \dots M \right]}$, such that the task is completed when each sub-goal state is reached. We denote a \textit{skill} $\pi(\cdot|l)$ as a reusable manipulation capability described by a language instruction $l$ (\eg `grasping', `placing', `folding') that can be instantiated across different objects and contexts \cite{brohan2022rt}. We make the assumption that for every sub-goal transition $g_i \to g_{i+1}$ in a task, there exists a reusable language-conditioned skill  $\pi(\cdot|l_{i+1})$ which can achieve the next desired sub-goal state (\eg a pick-and-place task would be decomposed into sub-goals \{`pick', `place'\}, with corresponding reusable skills to accomplish each transition). In this work, we are concerned with how to both learn representations of these sub-goals $g_i$, and learn language-conditioned skills $\pi(\cdot|l_i)$ from demonstrations, such that agents can generalize to novel task variations.

\textbf{Training data.} We assume access to robot teleoperation demonstrations $\mathcal{D}_{robot} = \{(o_t, \text{aux}_t, a_t, e_t, g)\}$ - workspace cameras $o_t$, auxiliary robot observations $\text{aux}_t$ (\eg wrist cameras, proprioception), actions $a_t$, end-effector poses $e_t$ and a language goal $g$, where $t$ denotes the timestep of the observation.

\textbf{Optional human data.} 
We optionally allow  an additional dataset of human videos $\mathcal{D}_{human} = \{(o_t, \hat{e}_t, g)\}$, where $\hat{e}_t$ is an estimated hand pose from an off-the-shelf tracker \cite{potamias2025wilor, ye2025predicting}. 
Human demonstrations do not provide action labels, so we use them only to train or adapt our high level policy $\pi_{hi}$ on estimated end-effector poses $\hat{e}_t$, keeping the low-level policy $\pi_{lo}$ trained purely on robot data.

\textbf{Assumptions.} We assume calibrated RGB-D workspace cameras for goal prediction, and that demonstrations contain sufficient visual cues to localize the hand near sub-goal boundaries. In addition, we assume that human demonstrations use similar grasp types as the robot gripper to minimize the embodiment gap. Full hardware configuration and per-task data collection details are provided in \suppref{B} and \suppref{C} respectively.

\section{GHOST} 
\label{sec:ghost}

\subsection{Overview}
The GHOST framework learns a hierarchy over \emph{sub-goal end-effector poses}. 
During training, sub-goal boundaries provide supervision for the high-level planner $\pi_{hi}$, while the full robot trajectories provide action supervision for the low-level controller $\pi_{lo}$. 
At test time, $\pi_{hi}$ predicts the next end-effector sub-goal (as a distribution) from workspace observations and language, and $\pi_{lo}$ executes goal-conditioned control toward that sub-goal.
This separation isolates long-horizon reasoning in $\pi_{hi}$ and maintains precise, embodiment-specific control in $\pi_{lo}$.

\subsection{Data Collection and Preprocessing}

\textbf{Robot Data.} We collect demonstrations through teleoperation across multiple
task variants, each demonstrating the same skill (\eg pick-and-place).
When possible, we extract sub-goals $s \in \mathcal{S}$ automatically by identifying the timesteps where the gripper state changes from open $\rightarrow$ close and vice versa.

\begin{figure}[h]
\centering
\includegraphics[width=\linewidth]{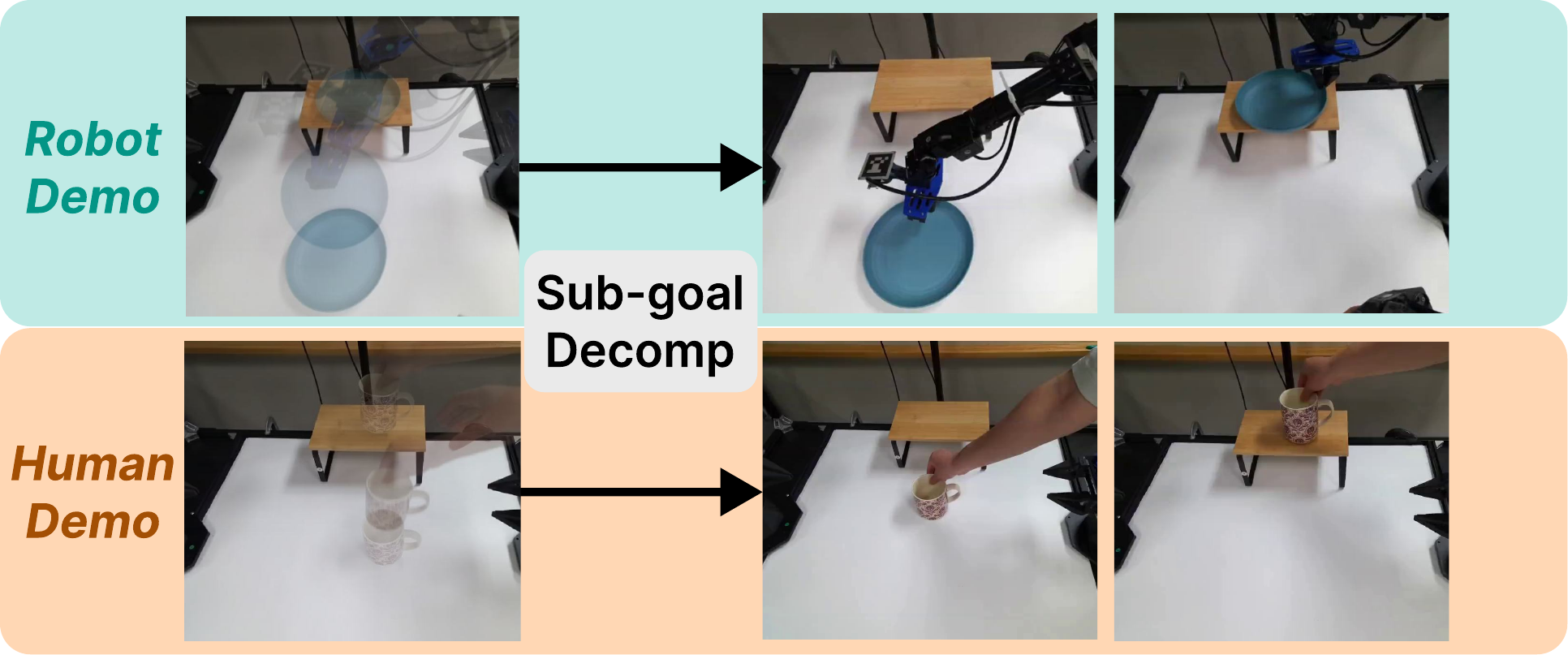}
\caption{Sub-goal decomposition for robot and human demonstrations. \textbf{Top}: Robot teleop demonstrations with sub-goals automatically extracted at gripper state transitions. \textbf{Bottom}: Human demonstrations with manually annotated sub-goals. Each sub-goal represents a semantically meaningful transition in the skill execution.}
\label{fig:sub-goal_decomp}
\end{figure}

\textbf{Human Data.} We optionally collect human demonstrations of the \textit{same skill} in a novel setting, using either new objects or novel applications of the skill. We track 3D hand poses $H_t$ using off-the-shelf hand pose estimators \cite{potamias2025wilor, ye2025predicting}. We resolve the weak-perspective scale ambiguity of the detection by segmenting the hand using Grounded-SAM \cite{ren2024grounded} and scaling the detection with the observed depth of the hand. We manually annotate sub-goals at the timesteps where a sub-goal terminates. We treat sub-goal boundaries as supervision rather than discovering them; we discuss automated discovery as future work in Sec.~\ref{sec:limitations}. Figure~\ref{fig:sub-goal_decomp} shows an example of our sub-goal decomposition for both robot and human demonstrations.

\begin{figure}[h]
\centering
\includegraphics[width=0.47\linewidth]{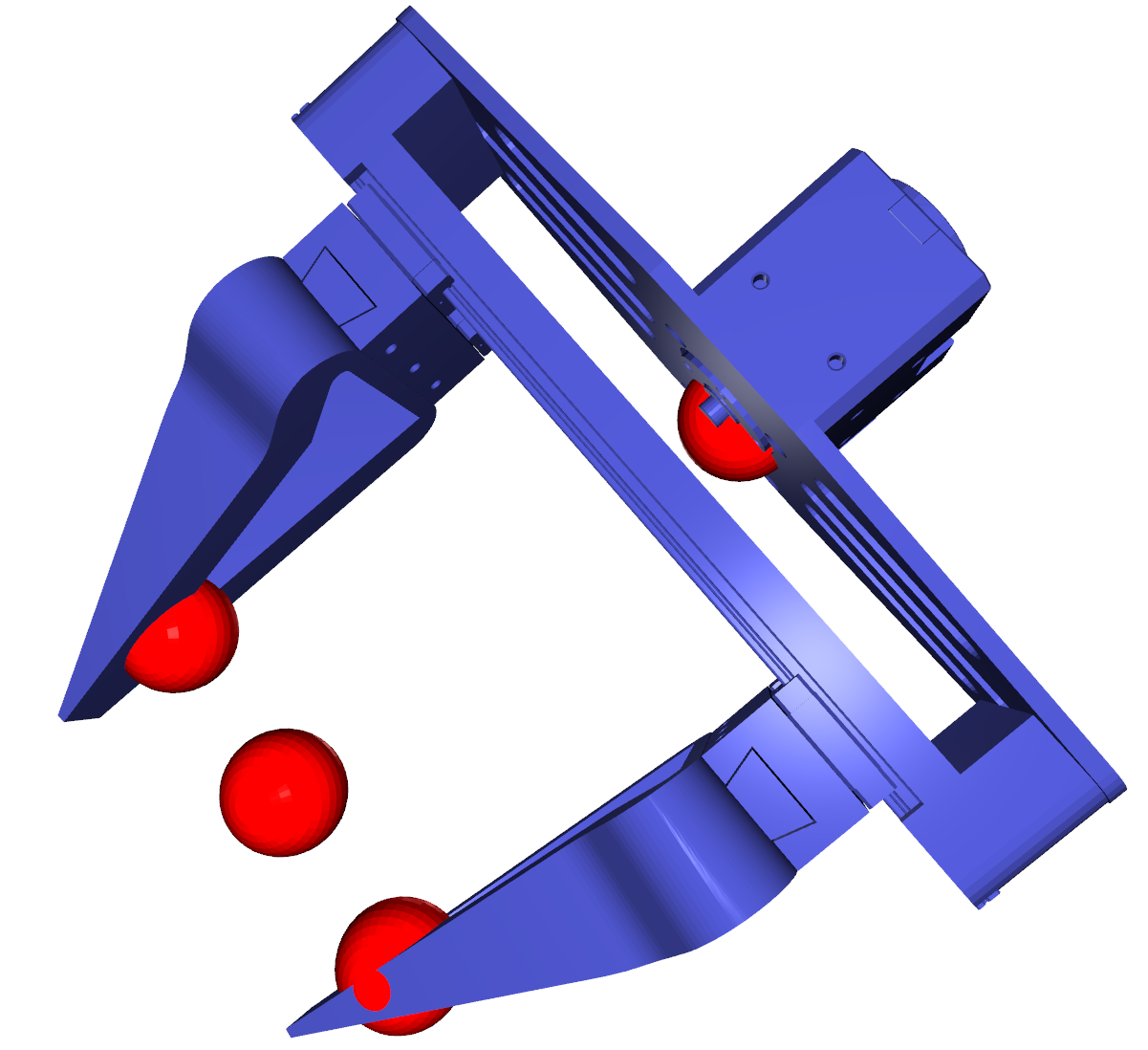}
\hfill
\includegraphics[width=0.47\linewidth]{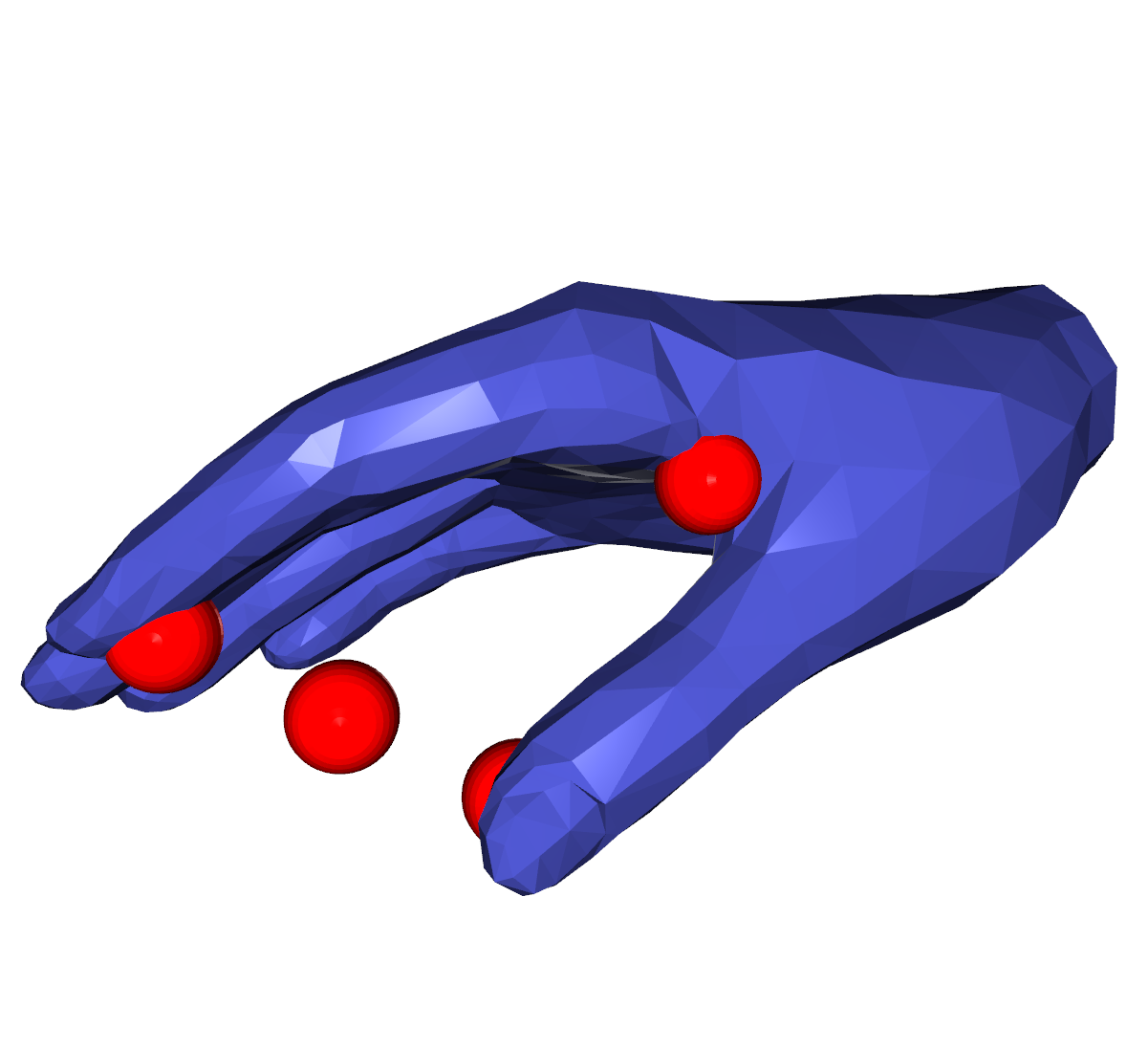}
\caption{End-effector representations for robot and human demonstrations. Instead of representing the gripper pose as a translation and SO(3) rotation, we represent the gripper pose as a set of 3D keypoints (red spheres): gripper base, fingertip locations, and grasping center. \textbf{Left}: Parallel-jaw gripper. \textbf{Right}: MANO hand.}
\label{fig:eef_repr}
\end{figure}

\textbf{Representing the end-effector pose}: Following prior work~\cite{wang2025articubot}, we represent the end-effector pose as a sparse set of 3D points instead of a position and $SO(3)$ rotation. For robot data, we sample a point cloud $P_R$ from the gripper mesh at each time-step and represent the end-effector pose $e_t$ as a set of 4 3D points from $P_R$, located at the base of the gripper, the tips of the two gripper fingers and the grasping center. Similarly, for human data, we extract hand point clouds $P_H$ from the MANO \cite{MANO:SIGGRAPHASIA:2017} mesh of the hand pose $H_t$ and select 4 3D points - corresponding to points on the palm, tips of the thumb and index finger, and the grasping center. Figure~\ref{fig:eef_repr} illustrates the keypoint locations for both gripper types. Thus, we extract a unified 3D end-effector representation for both human and robot data.

\begin{table*}[h]
\centering
\begin{tabular}{llcl}
\toprule
\textbf{Task} & \textbf{Data Source} & \textbf{\# Demos} & \textbf{Generalization Type} \\
\midrule
\multicolumn{4}{c}{\textit{Pick-and-Place}} \\
\midrule
\rowcolor{idblue!20} \texttt{plate-on-table} & Robot & 20 & — \\
\rowcolor{idblue!20} \texttt{plate-in-bin} & Robot & 20 & — \\
\rowcolor{idblue!20} \texttt{mug-in-bin} & Robot & 20 & — \\
\rowcolor{oodorange!20} \texttt{mug-on-table} & Human & 20 & Object combination \\
\midrule
\multicolumn{4}{c}{\textit{Cloth Folding}} \\
\midrule
\rowcolor{idblue!20} \texttt{fold-onesie} & Robot & 33 & — \\
\rowcolor{idblue!20} \texttt{fold-shirt} & Robot & 50 & — \\
\rowcolor{oodorange!20} \texttt{fold-onesie-ood} & Human & 17 & Object instance \\
\rowcolor{oodorange!20} \texttt{fold-towel} & Human & 50 & Object category + Skill composition \\
\midrule
\multicolumn{4}{c}{\textit{Hammer Pin}} \\
\midrule
\rowcolor{idblue!20} \texttt{hammer-pin} & Human+Robot & 100 & — \\
\midrule

\end{tabular}
\caption{Overview of tasks. \textcolor{idblue}{Blue}: In-distribution (ID) tasks. \textcolor{oodorange}{Orange}: Out-of-distribution (OOD) tasks requiring generalization.}
\label{tab:task_overview}
\end{table*}

\subsection{Policy Learning}

\subsubsection{High-Level Goal Prediction Policy}

The high-level policy $\pi_{hi}$ predicts a distribution over the end-effector pose $e_s$ 
at sub-goal timestep $s$.
We parameterize $\pi_{hi}$ as a decoder-only transformer \cite{jin2024lvsm} operating on RGB-D observations from $C$ cameras. The architecture is shown in Figure~\ref{fig:high-level-arch}.
Each RGB-D image is processed independently with a frozen DINOv3 \cite{siméoni2025dinov3} encoder, producing $K \times C$ patch tokens ($K$ per camera). Each patch token is augmented with the features of the 3D coordinate $[x,y,z]$ of the patch center, processed with an MLP. 
Additional tokens encode the current gripper pose  
and the task language embedding with Flan-T5~\cite{JMLR:v25:23-0870}. We also include a learnable token specifying whether the demonstration is from a human or robot.. All tokens are projected to the transformer's feature dimension through separate MLPs. We also add register tokens \cite{darcet2024visiontransformersneedregisters}, as they have been shown to improve performance on dense prediction tasks without any change in training objective. 

To handle the multimodality inherent in the demonstration data, we model the goal distributions as a dense per-patch Gaussian Mixture Model (GMM). Each patch token predicts: (1) a mixing weight $w_i$, and (2) four 3D residual vectors $\{\delta_{i,1}, \delta_{i,2}, \delta_{i,3}, \delta_{i,4}\}$ relative to the patch center $p_i$. These residual vectors are trained to predict the 3D points of the gripper. 
The global goal distribution is a mixture of $K \times C$ isotropic Gaussians (one per patch) with fixed variance $\sigma^2$\footnote{In practice we sum NLL losses over a small set of fixed variances to capture sub-goal distributions at multiple scales; see \suppref{A.1}.}. We do not claim the Dense GMM as a novel contribution in this paper, though we believe that this paper is the first to apply the idea of a Dense GMM in this per-patch manner (as opposed to using a Dense GMM over point clouds and  predicting a Gaussian per-point).

\begin{figure}[h]
\centering
\includegraphics[width=0.85\linewidth]{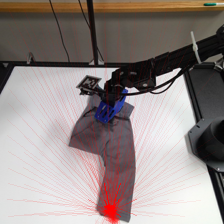}
\caption{High-level policy GMM predictions at inference. We visualize the mixture components of the GMM, with the opacity encoding mixing weight $w_i$ and arrows showing projections of 3D residuals $\delta_i$ from patch centers $p_i$.}
\label{fig:gmm_inference}
\end{figure}

\textbf{Training.} We train $\pi_{hi}$ by minimizing the negative log-likelihood on sub-goal transitions from a dataset of both human and robot data $\mathcal{D'} = \mathcal{D}_{robot} \cup \mathcal{D}_{human}$ that consists of observations $o_t$ for each timestep $t$ and end-effector poses $e_s$ at the end of each sub-goal $s$:
$$\mathcal{L}_{hi} = \mathbb{E}_{(o_t, e_s) \sim \mathcal{D'}} \left[-\log \left(\sum_{i=1}^{K \times C} w_i \mathcal{N}(e_s ; \mu_i, \sigma^2)\right)\right]$$
where $\mu_i$ is the predicted goal, given by the patch center + each of the residuals for patch $i$: $\mu_i = p_i + \delta_i$. Full architecture and training hyperparameters are listed in \suppref{A.1}.

\textbf{Inference.} During inference, we sample a patch $i$ from the categorical distribution on mixing weights $\{w_i\}$.  We  then use the corresponding mean prediction $\mu_i = p_i + \delta_i$
to obtain the predicted end-effector pose. Figure~\ref{fig:gmm_inference} visualizes the components of the GMM, depicting the spatial distribution over the predicted residuals for a single keypoint in a single camera view.

\subsubsection{Low-Level Goal Conditioned Policy}

We instantiate $\pi_{lo}$ as a Diffusion Policy \cite{chi2025diffusion}, which generates action chunks $\{a_t, ..., a_{t+H}\}$ by denoising conditioned on observations.

\begin{figure}[h]
\centering
\includegraphics[width=\linewidth]{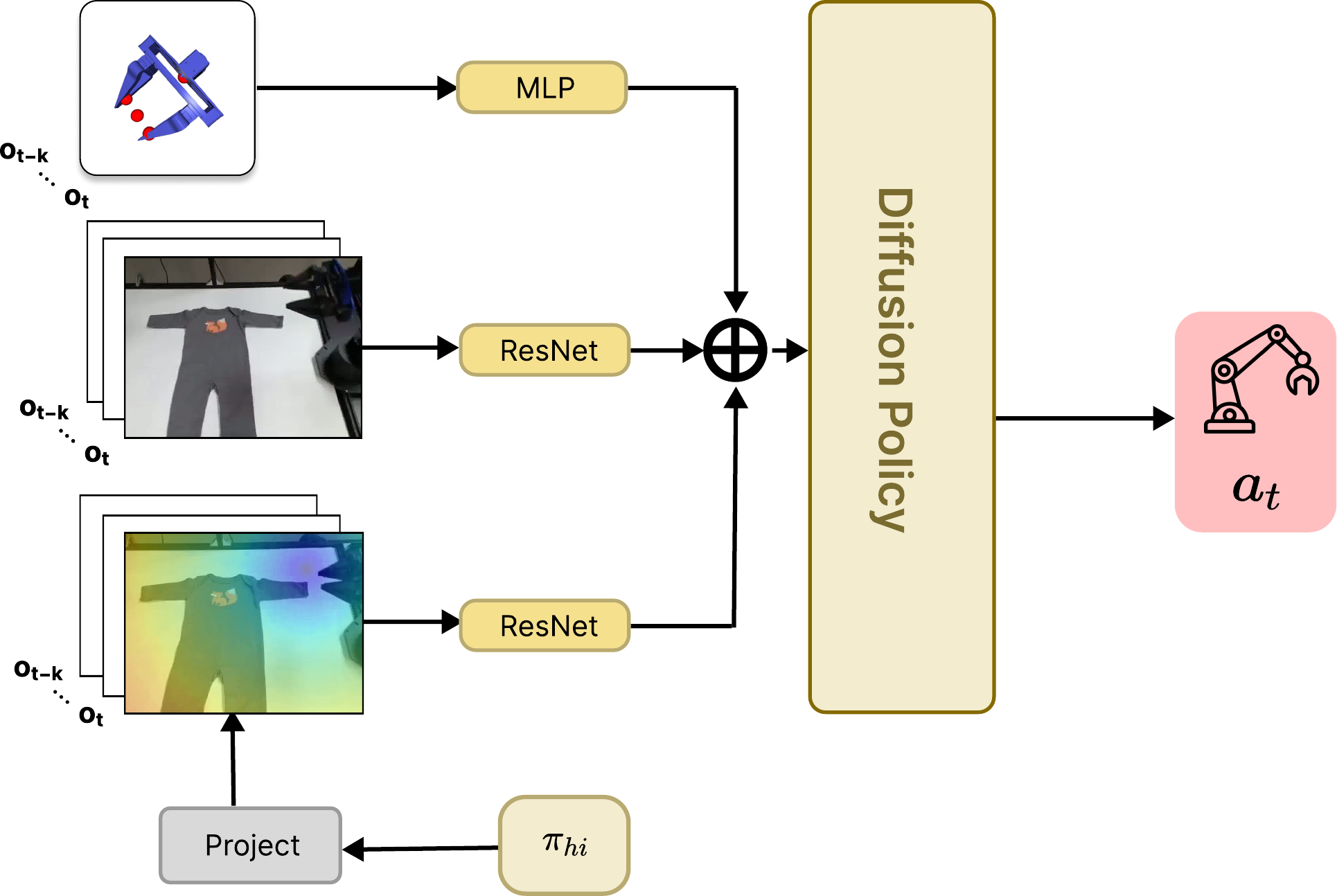}
\caption{\textbf{GHOST} Low-level goal-conditioned policy architecture. Images from each camera and the projected end-effector heatmap images are processed independently with ResNet \cite{he2016deep} encoders and concatenated along with the proprioceptive input into a global conditioning vector for the Diffusion Policy.}

\label{fig:low_level_arch}
\end{figure}

\begin{figure}[h]
\centering
\includegraphics[width=\linewidth]{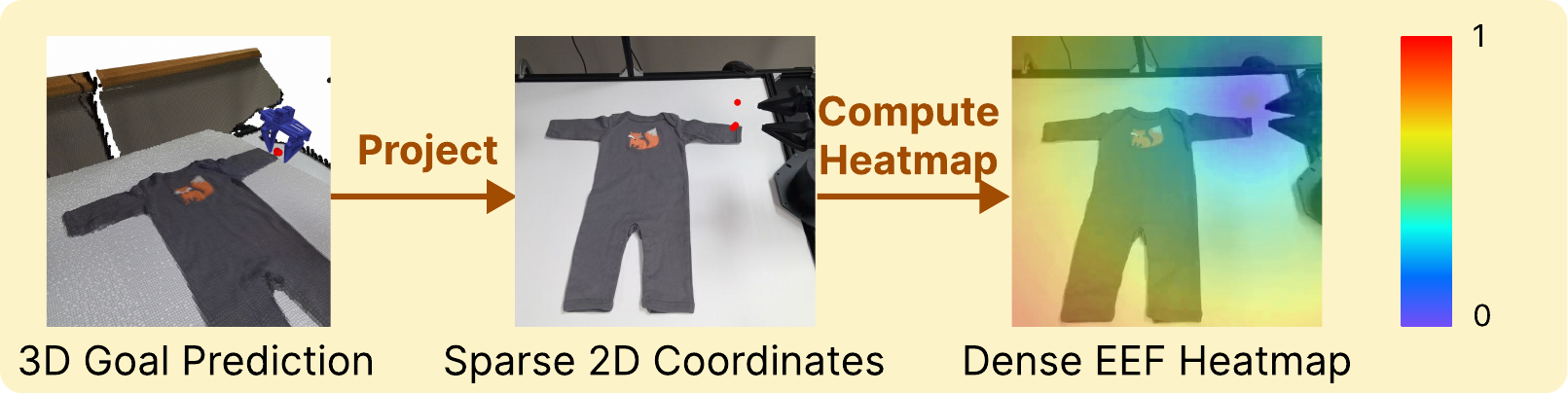}
\caption{End-effector heatmap generation for goal conditioning. Predicted 3D keypoints are projected to 2D coordinates, and then converted to dense distance field heatmaps that encode spatial proximity to each keypoint.}
\label{fig:eef_heatmap}
\end{figure}

\textbf{Goal Conditioning.} Each point in the 3D goal $e_s$ is projected onto the image plane of each camera using the camera parameters, yielding sparse 2D coordinates $\mathbf{p}_i \in \mathbb{R}^2$ for $i \in \{1\dots C\}$. These are converted to dense \textit{end-effector heatmaps} $H$ where each channel $c$ encodes the pixel distance field from keypoint $\mathbf{p}_c$:
$H_{xy}^{(c)} = \sqrt{\frac{\|\mathbf{x}_{xy} - \mathbf{p}_c\|_2}{d_{\text{max}}}}$
where $\mathbf{x}_{xy}$ denotes the pixel coordinates, $d_{\text{max}} = \sqrt{h^2 + w^2}$ normalizes by the image diagonal, and we use the square root for steeper gradients near the targets. We denote the resulting heatmap tensor as $\text{heatmap}(e_s) := H \in \mathbb{R}^{3 \times h \times w}$, where $\mathbf{x}_{xy}$ denotes the pixel coordinates. We use heatmaps rather than single-pixel binary masks as goal representations, empirically we find them to yield better performance; we hypothesize this is due to the denser supervisory signal. For practical implementation reasons, we select three of the heatmap channels corresponding to non-collinear points in the goal\footnote{We choose a point on the gripper wrist, as well as a point on each finger. Because of the rigid structure of the gripper, this minimal representation is sufficient to capture the state of the gripper, so discarding channels does not have a major effect on the policy.}, and pass these channels through standard 3-channel image encoder networks before incorporating into the policy network. This preserves spatial structure while densifying the sparse goal representation. This process is illustrated in Figure~\ref{fig:eef_heatmap}, and the architecture of the low-level goal conditioned policy is shown in Figure~\ref{fig:low_level_arch}.

\textbf{Training.} We use the standard diffusion loss, training on the robot-only data $\mathcal{D}_{robot}$:
$$\mathcal{L}_{lo} = \mathbb{E}_{\mathcal{D}_{robot}, \epsilon} \left[\|\epsilon - \epsilon_\theta(a_t^{\text{noisy}},o_t, \texttt{aux}_t, \text{heatmap}(e_s))\|^2\right]$$

where $\text{aux}_t$ are auxiliary robot observations  (\eg wrist cameras, proprioception), $a_t$ are low-level robot actions, $a_t^{\text{noisy}}$ denotes the action corrupted according to the forward diffusion process and $e_s$ is the gripper pose extracted from the robot demonstrations.
We train $\pi_{hi}$ and $\pi_{lo}$ independently. Full hyperparameters for $\pi_{lo}$ are listed in \suppref{A.2}.

\textbf{Inference.} At deployment, we execute $\pi_{hi}$ and $\pi_{lo}$ sequentially, where $\pi_{hi}$ predicts a sub-goal conditioned on the current observation and $\pi_{lo}$ predicts one action chunk conditioned on the heatmap representation of the sub-goal prediction.

\begin{table*}[h]
\centering
\begin{tabular}{lcc}
\toprule
 & \multicolumn{2}{c}{Success Rate (\%) $\uparrow$} \\
\cmidrule(lr){2-3}
Method & \textcolor{idblue}{\textbf{\texttt{plate-on-table}}} & \textcolor{oodorange}{\textbf{\texttt{mug-on-table}}} \\
\midrule
DP & 80.0 $\pm$ 15.0 & 13.3 $\pm$ 9.2 \\
MimicPlay & 65.0 $\pm$ 15.0 & 28.3 $\pm$ 12.5 \\
GHOST (Ours - Robot Only) & \underline{83.3 $\pm$ 10.8} & \underline{55.0 $\pm$ 15.0} \\
GHOST (Ours) & \colorbox{gray!25}{\textbf{98.3 $\pm$ 2.5}} & \colorbox{gray!25}{\textbf{63.3 $\pm$ 13.3}} \\
\bottomrule
\end{tabular}
\caption{Pick-and-place success rates. Best: \colorbox{gray!25}{\textbf{shaded}}, second-best: \underline{underline}.}
\label{tab:pick_place}
\end{table*}

\begin{table*}[h]
\centering
\begin{tabular}{llccccc}
\toprule
 & & \multicolumn{5}{c}{Success Rate (\%) $\uparrow$} \\
\cmidrule(lr){3-7}
Task & Method & 1-step & 2-step & 3-step & 4-step & Final \\
\midrule
\multirow{4}{*}{\textcolor{idblue}{\textbf{\texttt{fold-onesie}}}}
  & DP & 90.0 & \underline{76.7} & 53.3 & 40.0 & 10.0 $\pm$ 11.7 \\
  & MimicPlay & \underline{93.3} & \underline{76.7} & 70.0 & 70.0 & 46.7 $\pm$ 16.7 \\
  & GHOST (Ours - Robot Only) & \colorbox{gray!25}{\textbf{100.0}} & \colorbox{gray!25}{\textbf{100.0}} & \underline{96.7} & \underline{86.7} & \underline{80.0 $\pm$ 15.0} \\
  & GHOST (Ours) & \colorbox{gray!25}{\textbf{100.0}} & \colorbox{gray!25}{\textbf{100.0}} & \colorbox{gray!25}{\textbf{100.0}} & \colorbox{gray!25}{\textbf{90.0}} & \colorbox{gray!25}{\textbf{83.3 $\pm$ 13.3}} \\
\midrule
\multirow{4}{*}{\textcolor{oodorange}{\makecell[l]{\textbf{\texttt{fold-onesie-ood}}\\(Novel Object Instance)}}}
  & DP & \underline{76.7} & \underline{60.0} & 46.7 & 26.7 & 10.0 $\pm$ 10.0 \\
  & MimicPlay & 63.3 & 36.7 & 30.0 & 20.0 & 0.0 $\pm$ 0.0 \\
  & GHOST (Ours - Robot Only) & \colorbox{gray!25}{\textbf{100.0}} & \colorbox{gray!25}{\textbf{100.0}} & \underline{73.3} & \underline{60.0} & \underline{43.3 $\pm$ 16.7} \\
  & GHOST (Ours) & \colorbox{gray!25}{\textbf{100.0}} & \colorbox{gray!25}{\textbf{100.0}} & \colorbox{gray!25}{\textbf{93.3}} & \colorbox{gray!25}{\textbf{86.7}} & \colorbox{gray!25}{\textbf{56.7 $\pm$ 16.7}} \\
\bottomrule
\end{tabular}
\caption{Onesie-folding success rates. $k$-step = first $k$ folds completed. Final = all 5 folds completed. Best: \colorbox{gray!25}{\textbf{shaded}}, second-best: \underline{underline}.}
\label{tab:cloth_folding_onesie}
\end{table*}

\begin{table}[h]
\centering
\begin{tabular}{lc}
\toprule
Method & Success Rate (\%) $\uparrow$ \\
\midrule
MimicPlay & 16.7 $\pm$ 13.3 \\
GHOST (Ours) & \colorbox{gray!25}{\textbf{36.7 $\pm$ 16.7}} \\
\bottomrule
\end{tabular}
\caption{\textcolor{oodorange}{\textbf{\texttt{fold-towel}} (Novel Object Category + Skill Composition)} success rates. Best: \colorbox{gray!25}{\textbf{shaded}}.}
\label{tab:cloth_folding_towel}
\end{table}

\begin{table}[h]
\centering
\begin{tabular}{lc}
\toprule
Method & Success Rate (\%) $\uparrow$ \\
\midrule
DP & 16.7 $\pm$ 13.3 \\
MimicPlay & 33.3 $\pm$ 16.7 \\
GHOST (Ours - Robot Only) & \underline{50.0 $\pm$ 16.7} \\
GHOST (Ours) & \colorbox{gray!25}{\textbf{70.0 $\pm$ 16.7}} \\
\bottomrule
\end{tabular}
\caption{\textcolor{idblue}{\textbf{\texttt{hammer-pin}}} success rates. Best: \colorbox{gray!25}{\textbf{shaded}}, second-best: \underline{underline}.}
\label{tab:hammer_pin}
\end{table}

\begin{figure*}[htbp]
\centering
\includegraphics[width=0.8\linewidth]{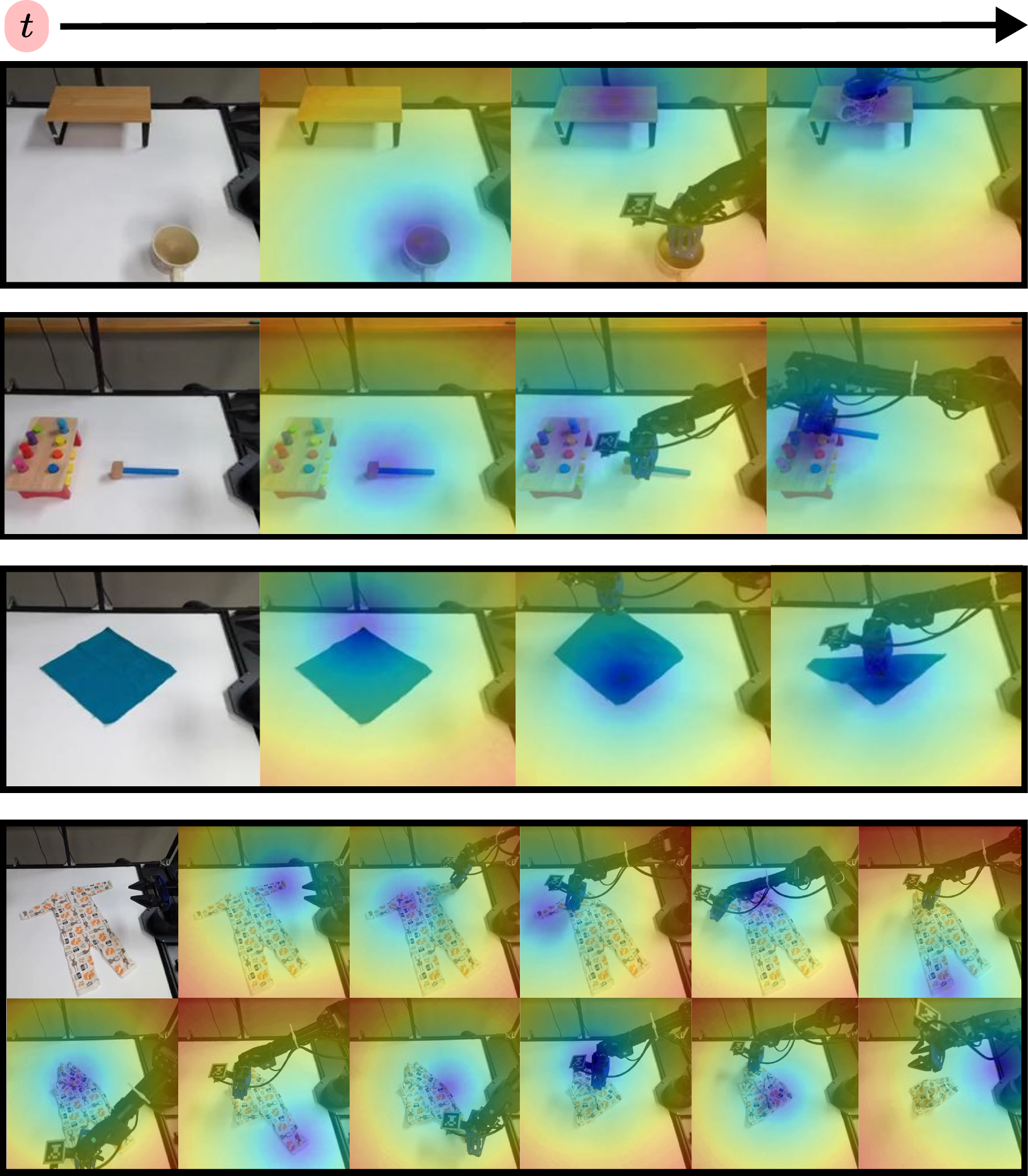}
\caption{We visualize qualitative results on various tasks with the \textbf{GHOST} framework. \textbf{Row 1}: \textcolor{oodorange}{\textbf{\texttt{mug-on-table}}} (novel object combination). \textbf{Row 2}: \textcolor{idblue}{\textbf{\texttt{hammer-pin}}}. \textbf{Row 3}: \textcolor{oodorange}{\textbf{\texttt{fold-towel}}} (novel category + skill composition). \textbf{Row 4}: \textcolor{oodorange}{\textbf{\texttt{fold-onesie-ood}}} (novel instance). We visualize rollouts by overlaying a colormap on the projection of $\pi_{hi}$'s goal predictions at each timestep.}
\label{fig:qualitative_results}
\end{figure*}

\section{Experiments and Results} 
\label{sec:experiments}

We evaluate whether our hierarchical policy learning approach improves performance and enables skill generalization.
Concretely, our experiments test two questions: \textbf{(Q1) In-distribution performance:} Do hierarchical policies improve in-distribution performance even without human data? \textbf{(Q2) Generalization:} Do human demonstrations enable transferring learned skills to novel object instances, categories, and contexts?

\subsection{Tasks and Evaluation Protocol}

\tbl{tab:task_overview} summarizes our evaluation tasks. We collect robot demonstrations on \textcolor{idblue}{in-distribution (ID)} tasks to train both $\pi_{hi}$ and $\pi_{lo}$, and human demonstrations on \textcolor{oodorange}{out-of-distribution (OOD)} variations to train only $\pi_{hi}$. For all experiments, we conduct $n=30$ trials per method with randomized object configurations, reporting mean success rate $\pm$ half-width of the 95\% percentile bootstrap CI (clipped to $[0,100]$).

\begin{itemize}
    \item \textbf{Pick-and-Place:} Grasp a specified object (plate, mug) and place on a target (table, bin). Robot demos include placing plates on a table/bin and placing a mug in the bin; human demos provide OOD (out of distribution) object-target combinations (see Table~\ref{tab:task_overview}). The OOD task \textcolor{oodorange}{\textbf{\texttt{mug-on-table}}} evaluates whether models can demonstrate \textit{compositional} generalization on a novel combination of objects. The task is considered a success if the object is placed stably on the target; objects placed successfully but knocked off during reset are given a score of $0.5$.
    \item \textbf{Cloth Folding:} Long-horizon deformable manipulation requiring multiple sequential folds. Robot demos on folding onesies and shirts; human demos on novel instances (\textcolor{oodorange}{\textbf{\texttt{fold-onesie-ood}}}) and applying the skill on entirely novel categories (\textcolor{oodorange}{\textbf{\texttt{fold-towel}}}). $k$-step success indicates first $k$ folds completed successfully.
    \item \textbf{Hammer Pin:} Pick up a hammer and strike a target pin. The blue, purple, red, and pink pins are initially raised. Robot demos include hammering the blue pin or the pink pin; human demos on novel target pin (purple pin). This task is considered a success if the target pin is fully driven in using the hammer.
\end{itemize}

\subsection{Baselines}

We evaluate GHOST against two baselines: (1) \textbf{Diffusion Policy (DP)} \cite{chi2025diffusion}, a flat policy trained on robot data, and (2) \textbf{MimicPlay} \cite{wang2023mimicplay}, a hierarchical approach that learns a latent trajectory representation from human play data. For fair comparison, we train our high-level policy architecture with MimicPlay's representation and training objective, appending a CLS token to the input sequence as the latent plan with a separate MLP as the GMM decoder. This ensures that we measure the differences that stem from the goal representation rather than policy architecture. We also evaluate \textbf{GHOST (Robot Only)}, our method trained without human demonstrations, to isolate the benefit of hierarchy from the benefit of human data.

\subsection{Implementation Details} 

We train a single multi-task policy with language conditioning for each method. Complete hyperparameters for both $\pi_{hi}$ and $\pi_{lo}$, our hardware setup, and per-task data collection details are provided in \suppref{A}, \suppref{B}, and \suppref{C} respectively.
For the Diffusion Policy baseline, we add language conditioning by appending the text embedding to the global conditioning vector. For hierarchical methods, language conditioning is only provided for the high-level policy.
We train all DP variants for 300k steps with color jitter and random crop augmentations. The high-level policy is trained for 100 epochs with additional augmentations: adding noise to the gripper pose, blur, grayscale, and token dropout augmentations. At inference, we synchronously execute $\pi_{hi}$ and $\pi_{lo}$, \ie the low-level predicts action chunks that fully execute before the next high-level inference step, enabling reactive replanning to changes in environment.

\subsection{Results}

\textbf{Do hierarchical policies improve in-distribution performance even without human data?} 
For \textcolor{idblue}{\textbf{\texttt{plate-on-table}}} (\tbl{tab:pick_place}) we see that nearly all methods saturate in performance, as the task is simple and sufficient training data is available. However, for a long-horizon complex task, we see a significant increase in the performance of ``GHOST (Ours - Robot Only)" compared to ``DP". As shown in \tbl{tab:cloth_folding_onesie}, on \textcolor{idblue}{\textbf{\texttt{fold-onesie}}} performance increases from 10\% to 80\% final success, showing large benefits from the hierarchical decomposition. Similarly, in \textcolor{idblue}{\textbf{\texttt{hammer-pin}}}, a task requiring precise grasping of the hammer tool and striking the correct pin, we see that performance significantly improves from DP (16.7\%) to ``GHOST (Ours - Robot Only)" (50\%) (\tbl{tab:hammer_pin}). \suppref{D} further isolates the contribution of the DINOv3 backbone via an ablation with a tiny ViT trained from scratch; the hierarchy alone (with no pre-trained encoder) already substantially outperforms flat DP, while DINOv3 provides additional gains on long-horizon tasks.

\textbf{Do human demonstrations enable transferring learned skills to novel object instances, categories, and contexts?}
As shown in \tbl{tab:pick_place}, \tbl{tab:cloth_folding_onesie} and \tbl{tab:cloth_folding_towel} human demonstrations unlock meaningful OOD transfer of learned skills to novel objects and novel skill compositions. GHOST achieves 63.3\% success on \textcolor{oodorange}{\textbf{\texttt{mug-on-table}}}, a task featuring a combination of objects unseen in robot demonstrations. On \textcolor{oodorange}{\textbf{\texttt{fold-onesie-ood}}}, GHOST achieves 56.7\% final success vs 43.3\% for ``GHOST (Ours - Robot Only)'' and 0\% for MimicPlay (\tbl{tab:cloth_folding_onesie}). Finally, on the hardest task of generalizing a policy to a novel object category and skill combination (\textcolor{oodorange}{\textbf{\texttt{fold-towel}}}), GHOST achieves 36.7\% success as compared to 16.7\% with the MimicPlay baseline.

\section{Limitations and Future Work} 
\label{sec:limitations}

Although GHOST shows significant gains in performance through a hierarchical framework, and promising results in generalizing learned skills across multiple axes, it is not without limitations. Like other methods that use hand pose estimators on human demonstrations \cite{lepert2025phantom, ren2025motion}, our approach is limited by the quality of the estimated hand pose, and suffers in situations where hand pose estimation has errors. We extract sub-goals through heuristics (gripper open/close), which may not be optimal or even valid for tasks requiring continuous manipulation skills (pouring, stirring, or hammering), and through annotation on human demonstrations, which imposes an additional cost on the collection of demonstrations. Related work on automatic sub-goal discovery \cite{pertsch2020long, zhang2024universal} is an interesting direction for future research on mining sub-goals from collected human and robot demonstrations. 

While GHOST generalizes to novel object instances and composes learned skills in novel sequences, we observe a steep drop in performance as the task horizon lengthens, which we attribute to the visual domain gap between human and robot observations in $\pi_{hi}$. To verify this, we conduct an oracle ablation (\suppref{E}): when $\pi_{hi}$ is trained on a small set of robot demonstrations of \textcolor{oodorange}{\textbf{\texttt{fold-towel}}} instead of human demonstrations---while $\pi_{lo}$ remains identical and has never seen towel-folding data---final success increases from 40\% to 90\%. This indicates that $\pi_{lo}$ generalizes zero-shot to novel object categories when given accurate sub-goals, and that the principal bottleneck for OOD generalization is the high-level visual domain gap rather than the low-level controller. Existing work on embodiment-invariant visual representations \cite{wang2023mimicplay, ren2025motion} aims to close precisely this gap and represents an important direction for future work.

\section{Conclusion} 
\label{sec:conclusion}

We present GHOST, a hierarchical imitation learning framework that decouples embodiment-agnostic goal prediction from embodiment-specific action execution. This factorization enables two key benefits: improved in-distribution performance through explicit sub-goal modeling, and out-of-distribution generalization via human demonstrations.

We introduce a 3D goal prediction architecture that processes multi-view RGB-D observations to predict dense per-patch Gaussian mixture models over end-effector poses. We condition the low-level policy through 2D projections of 3D goals, represented as end-effector heatmaps. This representation bridges the gap between human and robot demonstrations without requiring explicit action retargeting. By training a high-level policy on heterogeneous human and robot data and a low-level policy purely on robot demonstrations, we generalize learned skills across novel object-context combinations and compositional skill generalization.

\section*{Acknowledgments}

We would like to thank Alexis Hao, Mino Nakura, Kallol Saha and Pratik Bhowal for helpful discussions and assistance with data collection. We thank the members of the R-PAD lab for their feedback. This material is based on work supported by the Toyota Research Institute, the National Science Foundation under NSF CAREER Grant No. IIS-2046491, and an unrestricted gift from Google.

\bibliographystyle{plainnat}
\bibliography{references}

\clearpage
  \clearpage
\appendices

\section{Implementation Details}
\label{app:implementation}

\subsection{High-Level Policy}
\label{app:high_level}

The high-level policy uses a decoder-only transformer operating on tokens extracted from a frozen DINOv3~\cite{siméoni2025dinov3} backbone. Each image is randomly cropped and resized, with additional color jitter, Gaussian blur, and grayscale augmentations applied stochastically. We also randomly drop a subset of DINOv3 image tokens as a regularization strategy. Further, we also apply uniform noise to the gripper token to be robust to noisy estimates. 

After extracting patch tokens from DINOv3, we augment each token with a learned 3D positional embedding. Specifically, we unproject the center pixel of each patch of the depth image using camera parameters, and pass them through an MLP to produce the positional features.

The sub-goal distribution is parameterized as a dense GMM over all image patches. We train using multiple negative log-likelihood loss terms, with multiple fixed variances or using uniform mixing weights for the predicted distribution. Full hyperparameters are listed in \tbl{tab:dino3dgp_config}.

\subsection{Low-Level Policy}
\label{app:low_level}

We adopt Diffusion Policy~\cite{chi2025diffusion} (DP) as the low-level controller, using a ResNet-18 \cite{he2016deep} visual backbone trained from scratch with color jitter and random crop augmentations. The goal-conditioned variant receives the end-effector heatmap as input and does not use language conditioning. The baseline without goal-conditioning (\ie DP) encodes language instructions with SigLIP \cite{zhai2023sigmoid} and concatenates the resulting features to form the global conditioning vector. We use receding horizon control as in the original Diffusion Policy implementation with the hyperparameters described in \tbl{tab:diffusion_policy_config}.

\subsection{Hyperparameters}
\label{app:hyperparameters}

\begin{table}[H]
\caption{High-level policy hyperparameters.}
\label{tab:dino3dgp_config}
\centering
\begin{tabular}{ll}
\toprule
\textbf{Hyperparameter} & \textbf{Value} \\
\midrule
\multicolumn{2}{l}{\textbf{Transformer}} \\
Number of layers & 4 \\
3D Positional Features dim & 128 \\
Dropout & 0.1 \\
Number of register tokens & 4 \\
\midrule
\multicolumn{2}{l}{\textbf{Dense Per-Patch GMM}} \\
Fixed variances & \texttt{[0.01, 0.05, 0.1, 0.25, 0.5]} \\
Uniform weights coefficient & 0.1 \\
\midrule
\multicolumn{2}{l}{\textbf{Data Augmentation}} \\
Crop--resize probability & 0.3 \\
Crop size range & [192, 224] \\
Color jitter probability & 0.3 \\
Grayscale probability & 0.1 \\
Gaussian blur probability & 0.1 \\
\midrule
\multicolumn{2}{l}{\textbf{Training}} \\
Epochs & 100 \\
Learning Rate & $1 \times 10^{-4}$ \\
LR Scheduler & Cosine (100 warmup steps) \\
Optimizer & AdamW ($\beta_1{=}0.95$, $\beta_2{=}0.999$)\\
Batch Size & 128 \\
\bottomrule
\end{tabular}
\end{table}

\begin{table}[H]
\caption{Low-level policy hyperparameters.}
\label{tab:diffusion_policy_config}
\centering
\begin{tabular}{ll}
\toprule
\textbf{Hyperparameter} & \textbf{Value} \\
\midrule
\multicolumn{2}{l}{\textbf{Observation / Action}} \\
Observation horizon & 2 \\
Prediction horizon & 16 \\
Action horizon & 8 \\
\midrule
\multicolumn{2}{l}{\textbf{Vision Backbone}} \\
Architecture & ResNet-18 \\
Crop size & $700 \times 700$ \\
Crop jitter & 30 \\
Spatial softmax keypoints & 32 \\
\midrule
\multicolumn{2}{l}{\textbf{U-Net}} \\
Down dimensions & \texttt{[512, 1024, 2048]} \\
Kernel size & 5 \\
Number of groups & 8 \\
Diffusion step embedding dim & 128 \\
\midrule
\multicolumn{2}{l}{\textbf{Noise Scheduler}} \\
Type & DDPM \\
Training timesteps & 100 \\
Beta schedule & \texttt{squaredcos\_cap\_v2} \\
Prediction type & $\epsilon$ \\
Clip sample range & $[-1.0, 1.0]$ \\
\midrule
\multicolumn{2}{l}{\textbf{Training}} \\
Learning rate & $1 \times 10^{-4}$ \\
Optimizer & AdamW ($\beta_1{=}0.95$, $\beta_2{=}0.999$) \\
Weight decay & $1 \times 10^{-6}$ \\
LR scheduler & Cosine (500 warmup steps) \\
Batch Size & 4 \\
\bottomrule
\end{tabular}
\end{table}

\section{Hardware and Data Collection Setup}
\label{app:hardware}

We make use of the ALOHA \cite{zhao2023learning} robot platform, a bimanual robot consisting of WidowX arms for teleop (leader) and ViperX arms for manipulation (follower). However, we only make use of the right leader--follower pair; the left arm remains stationary throughout data collection and policy rollouts. All demonstrations are recorded at 30\,Hz from two Azure Kinect workspace cameras at a resolution of $1280 \times 720$ and then downsampled to 15\,Hz for policy learning. We also make use of an Intel Realsense wrist-mounted camera to augment the robot demonstrations. Rollouts are executed at 15\,Hz. Actions are represented as 10-dimensional absolute end-effector poses (3 position, 6 for rotation in the 6D representation \cite{zhou2019continuity} and one for gripper width) for the single active arm.

We use standard ArUco-marker-based extrinsics calibration for the workspace cameras, and apply photometric and geometric augmentations during training of both $\pi_{hi}$ and $\pi_{lo}$. We note that GHOST uses 3D information sparsely: $\pi_{hi}$ unprojects only the center pixel of each DINOv3 patch, 
and $\pi_{lo}$ conditions on heatmaps of a small set of projected keypoints rather than on dense point clouds. This sparse use of 3D reduces the surface area over which calibration or depth error can affect the policy compared to methods that consume dense point clouds or per-pixel 3D features.

\section{Task Descriptions}
\label{app:tasks}

\subsection{Pick-and-Place}

All pick-and-place tasks share a common workspace layout: the target object is placed at a randomized position on the table, while the receptacle (bin or toy table) is fixed at the center-back.

\begin{itemize}
    \item \textcolor{idblue}{\textbf{\texttt{mug-in-bin}}}: 3 mugs of distinct color and shape. 20 robot demonstrations distributed across mugs.
    \item \textcolor{idblue}{\textbf{\texttt{plate-on-table}}}: 3 plates of distinct color. 20 robot demonstrations distributed across plates.
    \item \textcolor{idblue}{\textbf{\texttt{plate-in-bin}}}: Same 3 plates as above. 20 robot demonstrations distributed across plates.
    \item \textcolor{oodorange}{\textbf{\texttt{mug-on-table}}}: Same 3 mugs as \texttt{mug-in-bin}. 20 human demonstrations distributed across mugs.
\end{itemize}

\subsection{Cloth Folding}

All cloth folding tasks place the garment at a random position and orientation on the table.

\begin{itemize}
    \item \textcolor{idblue}{\textbf{\texttt{fold-onesie}}}: 2 onesies of different color. 33 robot demonstrations distributed evenly.
    \item \textcolor{idblue}{\textbf{\texttt{fold-shirt}}}: 2 shirts of distinct color. 50 robot demonstrations distributed evenly.
    \item \textcolor{oodorange}{\textbf{\texttt{fold-onesie-ood}}}: A held-out onesie not seen during robot training. 17 human demonstrations.
    \item \textcolor{oodorange}{\textbf{\texttt{fold-towel}}}: 3 towels of distinct color. 50 human demonstrations distributed evenly.
\end{itemize}

\subsection{Hammer Pin}

\begin{itemize}
    \item \textcolor{idblue}{\textbf{\texttt{hammer-pin}}}: The hammer is placed at a random position; the pin board is placed near the table center. We collect robot demonstrations of striking the blue and pink pins, and human demonstrations of striking purple and blue+purple pins. We test with the blue pin for evaluation. 25 demonstrations for each of the 4 datasets.
\end{itemize}

\section{Ablation: Visual Backbone for High-Level Policy}
\label{app:visual_backbone}

We ablate the choice of visual backbone by replacing the frozen DINOv3 encoder in $\pi_{hi}$ with a tiny ViT (11M parameters, comparable to ResNet-18) trained from scratch. This isolates whether GHOST's gains stem from the pre-trained visual encoder or from the hierarchical factorization itself. We evaluate on \textcolor{idblue}{\textbf{\texttt{fold-onesie}}}, our most demanding in-distribution task.

\begin{table}[H]
\centering
\begin{tabular}{lccccc}
\toprule
 & \multicolumn{5}{c}{Success Rate (\%) $\uparrow$} \\
\cmidrule(lr){2-6}
Method & 1-step & 2-step & 3-step & 4-step & Final \\
\midrule
DP & \underline{80} & \underline{60} & 50 & 40 & 20 \\
GHOST (ViT) & \colorbox{gray!25}{\textbf{100}} & \colorbox{gray!25}{\textbf{100}} & \underline{90} & \underline{70} & \underline{50} \\
GHOST (DINOv3) & \colorbox{gray!25}{\textbf{100}} & \colorbox{gray!25}{\textbf{100}} & \colorbox{gray!25}{\textbf{90}} & \colorbox{gray!25}{\textbf{90}} & \colorbox{gray!25}{\textbf{100}} \\
\bottomrule
\end{tabular}
\caption{Ablation on the high-level visual backbone for \textcolor{idblue}{\textbf{\texttt{fold-onesie}}}. GHOST (ViT) uses a tiny ViT trained from scratch in place of the frozen DINOv3 encoder. Best: \colorbox{gray!25}{\textbf{shaded}}, second-best: \underline{underline}. $n=10$ rollouts.}
\label{tab:vit_ablation}
\end{table}

As shown in \tbl{tab:vit_ablation}, GHOST with a ViT trained from scratch (50\% final success) still substantially outperforms the flat Diffusion Policy baseline (20\%), confirming that the hierarchical sub-goal factorization is itself a significant contributor to performance. However, the frozen DINOv3 backbone provides a further large improvement (90\%), indicating that a strong pre-trained visual encoder is important for accurate goal prediction, particularly over longer horizons where errors in $\pi_{hi}$ compound. Thus, both the hierarchy and the visual backbone contribute meaningfully to GHOST's performance.

\section{Ablation: Oracle High-Level Policy}
\label{app:oracle_high_level}

To disentangle the contribution of goal prediction quality (from human demonstrations) and goal-following ability, we collect an additional 25 robot teleoperation demonstrations of \textcolor{oodorange}{\textbf{\texttt{fold-towel}}} and train the high-level policy on this data instead of human demonstrations. This gives us an ``oracle'' $\pi_{hi}$ that operates without any embodiment gap, while $\pi_{lo}$ remains identical to all other experiments---crucially, it has never seen \textcolor{oodorange}{\textbf{\texttt{fold-towel}}} during training.

\begin{table}[H]
\centering
\begin{tabular}{lc}
\toprule
Method & Success Rate (\%) $\uparrow$ \\
\midrule
MimicPlay & 20 \\
GHOST (human $\pi_{hi}$) & 40 \\
GHOST (oracle $\pi_{hi}$) & \textbf{90} \\
\bottomrule
\end{tabular}
\caption{Oracle high-level ablation on \textcolor{oodorange}{\textbf{\texttt{fold-towel}}}. The oracle $\pi_{hi}$ is trained on robot demonstrations; $\pi_{lo}$ is unchanged and has never seen towel-folding data.}
\label{tab:oracle_high_level}
\end{table}

Replacing human demonstrations with robot data in $\pi_{hi}$ increases success from 40\% to 90\%, while $\pi_{lo}$ is held fixed. This reveals two things. First, the low-level policy can generalize zero-shot to a novel object category (towels) when given accurate sub-goals, confirming that the goal-conditioned factorization enables meaningful skill transfer. Second, the primary bottleneck for OOD generalization is the visual domain gap between human and robot observations in $\pi_{hi}$, rather than the low-level controller's ability to execute on unseen objects. Closing this embodiment gap in the high-level policy is an important direction for future work.

\section{Computational Cost}
\label{app:compute}
We report wall-clock inference time for each method, averaged over 5 rollouts on a single NVIDIA RTX 4090 GPU. All methods predict an action chunk of length 16.
\begin{table}[H]
\centering
\begin{tabular}{lc}
\toprule
\textbf{Method} & \textbf{Inference time (ms / step)} \\
\midrule
Diffusion Policy & $60.1 \pm 0.8$ \\
MimicPlay & $127.8 \pm 1.2$ \\
GHOST ($\pi_{hi} + \pi_{lo}$) & $149.3 \pm 2.1$ \\
\bottomrule
\end{tabular}
\caption{Per-step inference time (mean $\pm$ std) on a single RTX 4090. }
\label{tab:inference_time}
\end{table}

\end{document}